\documentclass[letterpaper, 10 pt, conference]{ieeeconf}  

\IEEEoverridecommandlockouts                              

\usepackage{mathtools}
\usepackage{hyperref}
\usepackage{bm} 
\usepackage{amssymb}
\usepackage{amsfonts}
\usepackage{mathbbol} 
\usepackage{graphicx, import} 
\usepackage{float}
\usepackage{times}
\usepackage{multicol}
\usepackage{multirow}    
\usepackage{microtype}
\usepackage[all]{nowidow}
\usepackage{cases}
\usepackage{yfonts}
\usepackage{enumerate}   
\usepackage{subcaption}
\usepackage{subcaption}
\usepackage{colortbl}
\usepackage{xparse}
\usepackage{arydshln}
\usepackage{microtype}
\usepackage{soul}
\usepackage[skins]{tcolorbox}
\usepackage{xcolor}
\usepackage{tikz}
\usepackage[mediumspace,mediumqspace,squaren]{SIunits}
\usepackage{svg}

\usepackage{tabularx,booktabs}
\newtcolorbox{myframe}[2][]{%
  enhanced,colback=white,colframe=black,coltitle=black,
  sharp corners,boxrule=0.4pt,left=0pt,right=0pt,top=0pt,bottom=0pt,
  fonttitle=\itshape,
  attach boxed title to top left={yshift=-0.3\baselineskip-0.4pt,xshift=2mm},
  boxed title style={tile,size=minimal,left=0.5mm,right=0.5mm,
  colback=white,before upper=\strut},
  title=#2,#1
}
  
\DeclareMathOperator*{\minimize}{minimize}

\DeclareMathOperator*{\argmax}{argmax}

\title{\LARGE \bf Automatic Gain Tuning for Humanoid Robots Walking Architectures Using Gradient-Free Optimization Techniques}
\author{Carlotta Sartore$^{*,1,2}$, Marco Rando$^{*,3}$, Giulio Romualdi$^{1}$, Cesare Molinari$^{4}$, Lorenzo Rosasco$^{3,5}$, Daniele Pucci$^{1,2}$
\thanks{$^{1}$Artificial and Mechanical Intelligence Istituto Italiano di Tecnologia,
Center for Robotics Technologies, Genova, Italy.
        {\tt\small name.surname@iit.it}}%
\thanks{$^{2}$School of Computer Science, The University of Manchester,
Manchester, United Kingdom.}%
\thanks{$^3$MaLGa-DIBRIS, University of Genoa, Italy.}
\thanks{$^4$MalGa-DIMA, University of Genoa, Italy.}
\thanks{$^5$CBMM-MIT, Cambridge, MA, USA.}
\thanks{$^*$ These authors contributed equally to this work.}
}
\begin{document}

\maketitle
\thispagestyle{empty}
\pagestyle{empty}


\begin{abstract}
Developing sophisticated control architectures has endowed robots, particularly humanoid robots, with numerous capabilities. However, tuning these architectures remains a challenging and time-consuming task that requires expert intervention. 
In this work, we propose a methodology to automatically tune the gains of all layers of a hierarchical control architecture for walking humanoids. We tested our methodology by employing different gradient-free optimization methods: Genetic Algorithm (GA), Covariance Matrix Adaptation Evolution Strategy (CMA-ES), Evolution Strategy (ES), and Differential Evolution (DE). We validated the parameter found both in simulation and on the real ergoCub humanoid robot. Our results show that  GA achieves the fastest convergence ($10\times 10^3$ function evaluations vs $25 \times 10^3$ needed by the other algorithms) and $100$\% success rate in completing the task both in simulation and when transferred on the real robotic platform.  These findings highlight the potential of our proposed method to automate the tuning process, reducing the need for manual intervention. 
\end{abstract}
\section{INTRODUCTION}
\label{sec:introduction}
 In recent years, humanoid robots, have gained interest as their range of tasks and capabilities continuously expand \cite{haarnoja2024learning,kim2021bipedal,amatucci2024vero}. This progress is also due to the development of sophisticated control architectures, which are becoming increasingly complex. 

\begin{figure}
\centering
\begin{subfigure}{0.32\columnwidth}
    \centering
    \includegraphics[trim=37.9cm 7.0cm 37.5cm 12.0cm, clip=true, width=\columnwidth]{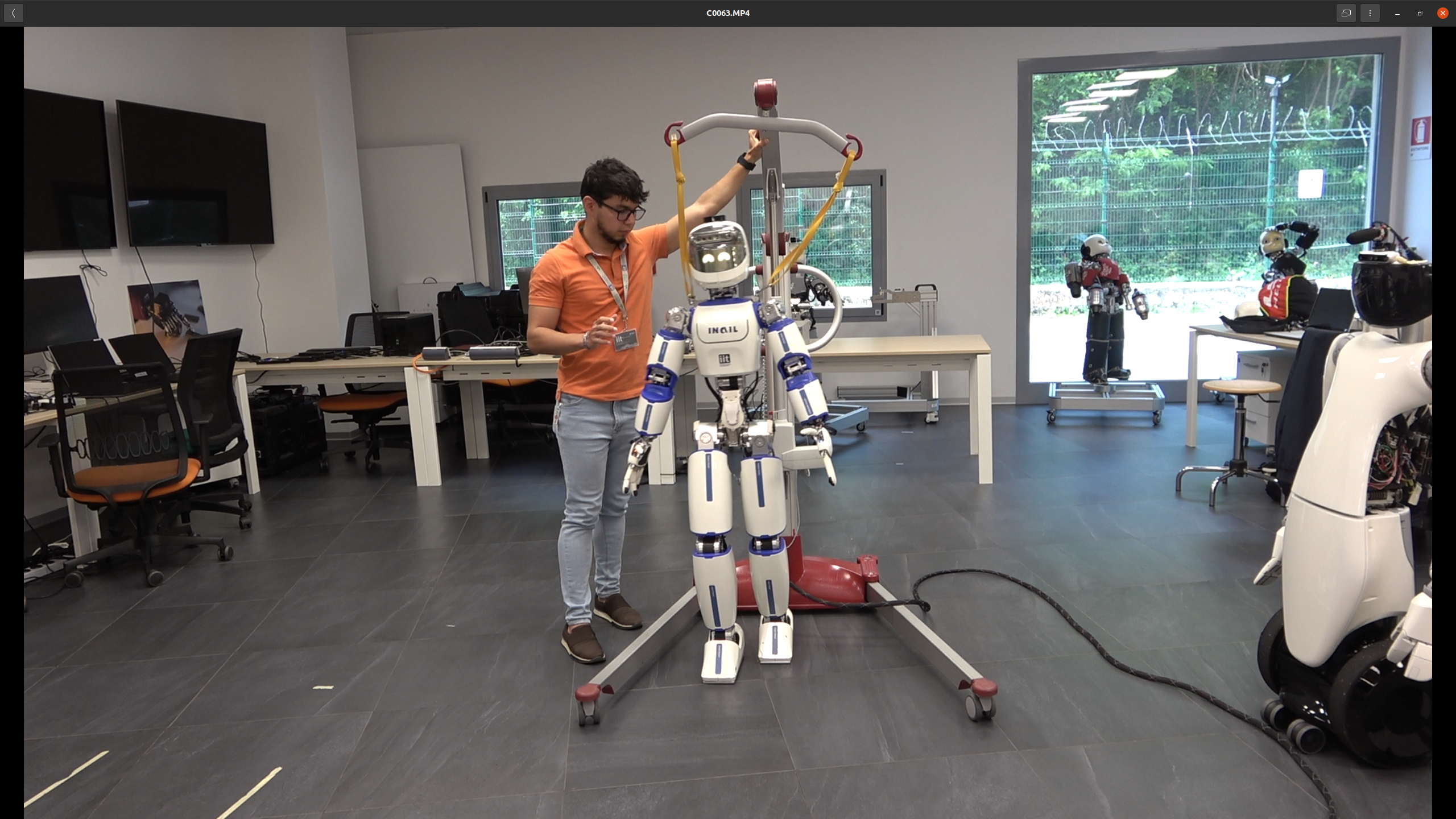}
\end{subfigure}
\hfill
\begin{subfigure}{0.32 \columnwidth}
    \centering
    \includegraphics[trim=36.4cm 7.0cm 39.cm 12.0cm, clip=true, width=\columnwidth]{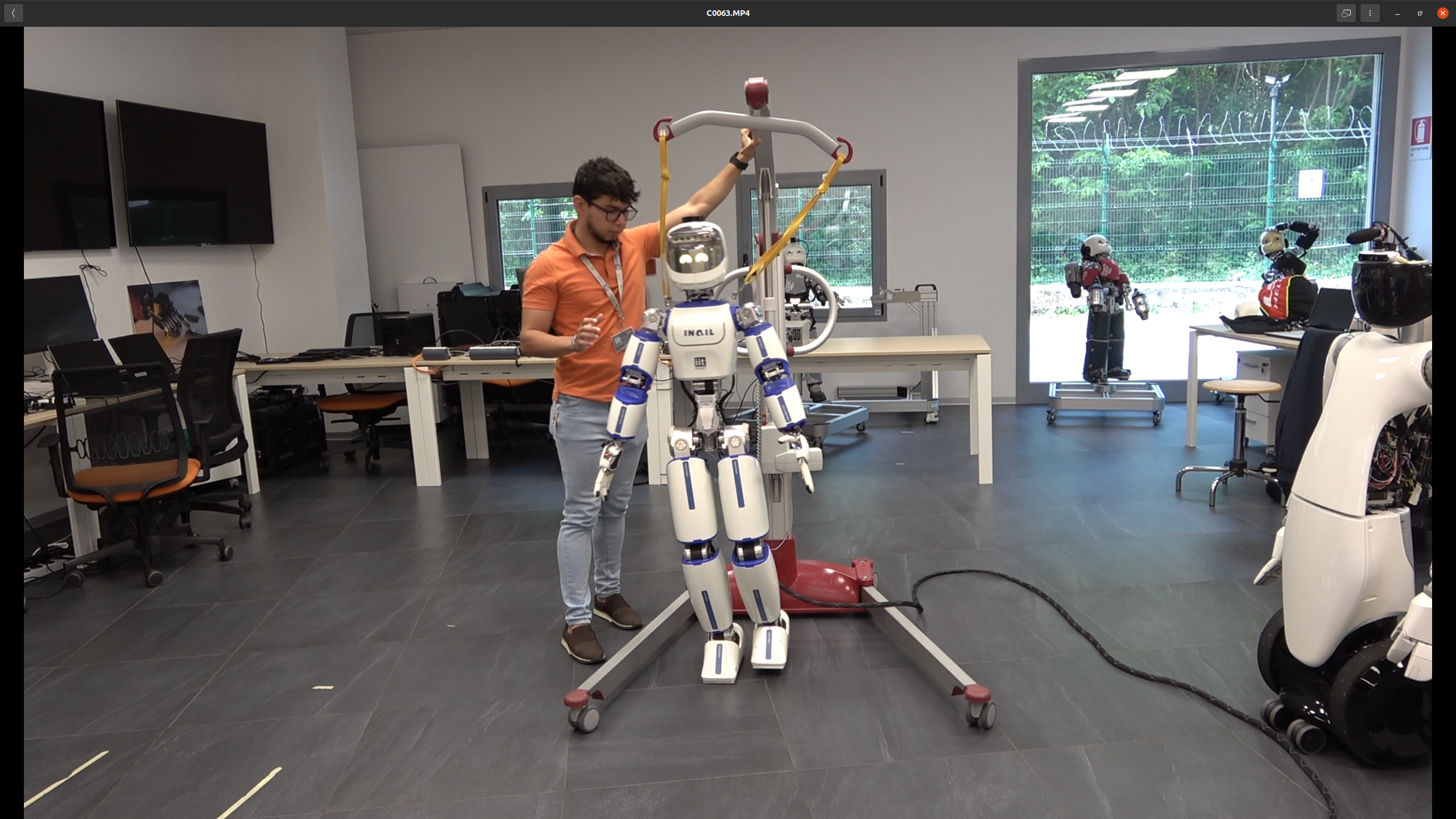}
\end{subfigure}
\hfill
\begin{subfigure}{0.32 \columnwidth}
    \centering
    \includegraphics[trim=40.0cm 7.0cm 35.4cm 12.0cm, clip=true, width=\columnwidth]{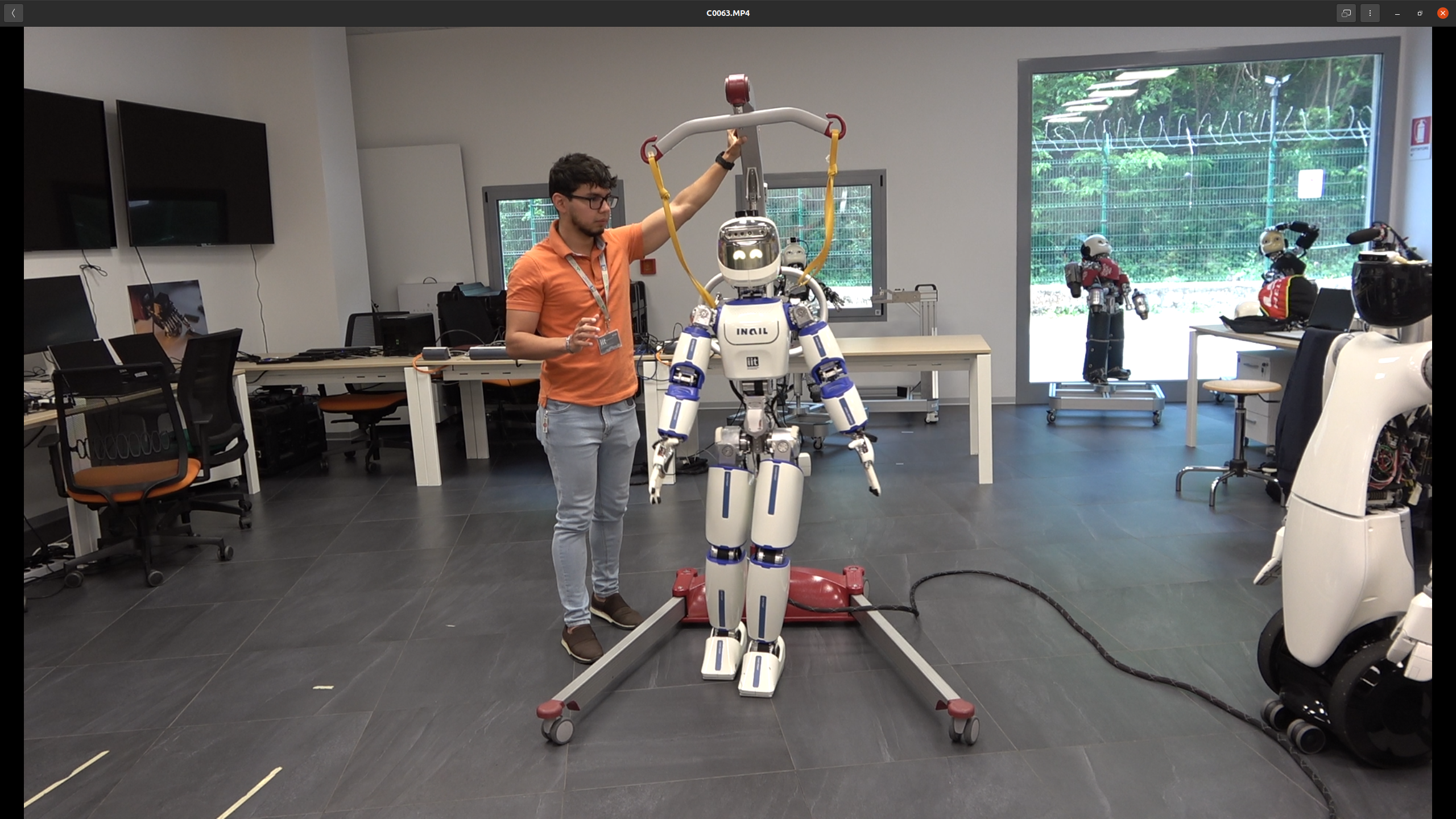}
\end{subfigure}
\hfill
\caption{The ergoCub robot walking with an optimized control architecture, defined by parameters identified through gradient-free techniques. } 
\label{fig:ergoCub_walking}
\end{figure}

When defining control architectures for legged robots, two main approaches are predominant in the literature: \emph{Reinforcement Learning (RL)} \cite{sutton1999reinforcement} and \emph{model based hierarchical control architecture} \cite{feng2015optimization}. RL has proven to be highly effective in enabling legged robots to perform a wide variety of tasks \cite{tsounis2020deepgait,lee2020learning}, demonstrating great capabilities in executing agile movements. However, even though such techniques show promising results, they are data-demanding and challenging to port on real robotic platforms since they lack theoretical stability guarantees \cite{shi2024rethinking}. On the other hand, model-based classical hierarchical control architectures come with theoretical guarantees  \cite{Galloway_2015,Pandala_2024} and they have been widely employed to equip robots with locomotion capabilities \cite{dantec2022whole} and agile maneuvers \cite{Chignoli2020}. Anyhow, tuning the numerous parameters of these architectures remains a tedious and time-consuming process that necessitates expert intervention.
For these reasons, several works have proposed automatic gain tuning of such architectures using various optimization techniques. \emph{Bayesian Optimization (BO)} \cite{berkenkamp2023bayesian} has been widely employed in the literature to tune the parameters of classical control architectures automatically. In \cite{Yang2022,Berkenkamp2016}, constrained BO is used to tune gains both in simulation and on real robotic platforms while adhering to safety constraints. However, BO performance deteriorates as the search space dimension increases \cite{calandra2016bayesian}, thus limiting its application to tuning only a limited part of the control architecture. Moreover, in \cite{Berkenkamp2016} an initial safe parameter configuration must be provided. In \cite{Rai2018}, the search space of BO optimization was increased using domain knowledge, but this required a search space transformation based on physiotherapist metrics. In \cite{Schperberg2022}, the author proposed the use of an \emph{Unscented Kalman Filter} to tune online the gains and weights of a swing and stance controller to satisfy user-specific needs. This approach was validated in simulation on a quadruped robot, showing fast convergence and avoiding the need for a trial-and-error setup. However, this interesting method has yet to be validated on a real robotic platform and applied to a humanoid robot. Furthermore, control architectures and RL have been used together in various studies. For instance, in \cite{Song2022}, Model Predictive Control (MPC) decision variables are learned using RL. However, the proposed method is computationally expensive and data-demanding. Additionally, RL is used only to tune a subpart of the control architecture, specifically the MPC parameters. 

\noindent In this work, we introduce a methodology to automatically tune the gains of all the layers composing the cascade walking control architecture and we compare the performances of several gradient-free optimization techniques in solving the task. Our contributions are as follows:

\begin{enumerate}[(i)]
\item We propose a methodology to tune all layers of the control architecture.
\item We compare four gradient-free techniques, namely Genetic Algorithm (GA), Covariance Matrix Adaptation Evolution Strategy (CMA-ES), Evolution Strategy (ES) and Differential Evolution (DE).
\item We validate the obtained results on the real ergoCub robotic platform using a reference trajectory different from the one utilized during the optimization. 
\end{enumerate}

\noindent Our results show that the proposed methodology successfully optimizes the architecture parameters with respect to the defined objective function, both in simulation and on the real robot.
Furthermore, the results indicate that among the gradient-free methods analyzed, the GA exhibits the fastest convergence and best performance, with $100$ \% success rate both in simulation and on the real robot. 

\noindent The rest of the paper is organized as follows: Sec. \ref{sec:background} presents the modeling and hierarchical control architecture layers used. Sec. \ref{sec:gain_optimization} formulates the automatic gain-tuning optimization problem. Sec. \ref{sec:results} presents the optimization and validation results. Finally, Sec. \ref{sec:conclusions} draws conclusions and highlights possible directions for improvement.

\section{BACKGROUND}
\label{sec:background}
\subsection{Notation}
\label{sec:background:notation}
\begin{itemize}
    \item $\mathcal{I}$ denotes the inertial frame of reference.
    \item  $\prescript{\mathcal{A}}{}{p}_{\mathcal{B}} \in \mathbb{R}^3$ is the the position of the origin of the frame $\mathcal{B}$ with respect to the frame $\mathcal{A}$.
    \item  $\prescript{\mathcal{A}}{}{R}_{\mathcal{B}} \in SO(3)$ represents the rotation matrix of the frame $\mathcal{B}$ with respect to $\mathcal{A}$.
    \item $\prescript{\mathcal{A}}{}{\omega}_{\mathcal{B}} \in \mathbb{R}^3$ is the angular velocity of the frame $\mathcal{B}$ with respect to $\mathcal{A}$, expressed in $\mathcal{A}$.
    \item The operator $\text{sk}(.) :\mathbb{R}^{3 \times 3} \to SO(3)$ denotes \textit{skew-symmetric} operation of a matrix, such that given $A \in \mathbb{R}^{3 \times 3}$, it is defined as $\text{sk}(A) := (A - A^\top)/2$.
    \item The \textit{vee} operator $^{\vee} : SO(3) \to \mathbb{R}^{3}$ denotes the inverse of \textit{skew-symmetric} vector operator. Given $A \in SO(3)$, $A^{\vee}\in \mathbb{R}^3$ is the vector such that $A^{\vee} \times u=Au$  for every $u \in \mathbb{R}^{3}$.
     \item The operator $\left\lVert . \right\rVert_W$ indicates the norm weighted by $W$. 
     \item $g$ is the gravity vector expressed in  $\mathcal{I}$.
    \item $\mathbb{I}_3$ and $0_3$ are the identity and zero matrices of dimension 3, respectively.
\item The force acting on a point of a rigid body is uniquely identified by the wrench $\prescript{}{}{\mathrm{f}}_\mathcal{B}^ \top = \begin{bmatrix} \prescript{\mathcal{A}}{}{{f}}_\mathcal{B} ^ \top & \prescript{\mathcal{A}}{}{\mu}_\mathcal{B}^ \top \end{bmatrix}$, where $\prescript{\mathcal{A}}{}{{f}}_\mathcal{B} \in \mathbb{R}^3$ denotes the force acting on the rigid body attached to the frame $\mathcal{B}$ expressed in $\mathcal{A}$. $\prescript{\mathcal{A}}{}{\mu}_\mathcal{B} \in \mathbb{R}^3$ denotes the moment of a force about the origin of $\mathcal{B}$ expressed in $\mathcal{A}$.
\item Whenever the superscripts are dropped, quantities are referred to the inertial frame.
\end{itemize}
\subsection{Modelling}
\noindent A humanoid robot is a multi-body mechanical system composed of $n+1$ rigid \textit{links} connected by $n$ \textit{joints}. None of the links have a prior constant \emph{pose}, hence position and orientation, with respect to the inertial reference frame. We refer to this system as a \textit{floating base}, where the so-called \textit{base-frame }, denoted with $\mathcal{B}$, is attached to a specific link of the system.  The \textit{model configuration} is defined as $q = ({p}_{\mathcal{B}}, {R}_{\mathcal{B}}, s)\in \mathbb{Q}=\mathbb{R}^{3} \times SO(3) \times \mathbb{R}^{n}$, where ${p}_{\mathcal{B}}$ and ${R}_{\mathcal{B}}$ denote respectively the position and the orientation of the \textit{base frame}, and $s$ is the joints configuration.
The \textit{model velocity} is $\nu=({\mathrm{v}}_{\mathcal{B}}, \dot{s}) \in \mathbb{V}= \mathbb{R}^{6+n}$, where ${\mathrm{v}}_{\mathcal{B}}=({\dot{p}}_{\mathcal{B}},{\omega}_{\mathcal{B}}) \in \mathbb{R}^6$ denotes the linear and angular velocity of the \textit{base frame}, and $\dot{s}$ denotes the joint velocities. 
Given a frame $\mathcal{A}$ rigidly attached to the kinematic chain, it is possible to obtain its pose via a geometrical forward kinematics map $h_{\mathcal{A}}(\cdot):\mathbb{Q} \to SO(3) \times \mathbb{R}^3$, while the map from the system velocity $\nu$ to the frame velocity ${\mathrm{v}}_{\mathcal{A}}$ is obtained via the \textit{Jacobian} ${J}_{\mathcal{A}}={J}_{\mathcal{A}}(q) \in \mathbb{R}^{6 \times (n+6)}$, i.e. ${\mathrm{v}}_{\mathcal{A}} = {J}_{\mathcal{A}}(q)  \nu$. 
\subsection{Walking Hierarchical Control Architecture}
\begin{figure*}
\centering
\includegraphics[trim=0.0cm 0.9cm 2.2cm 0.0cm, clip=true, scale=.9]{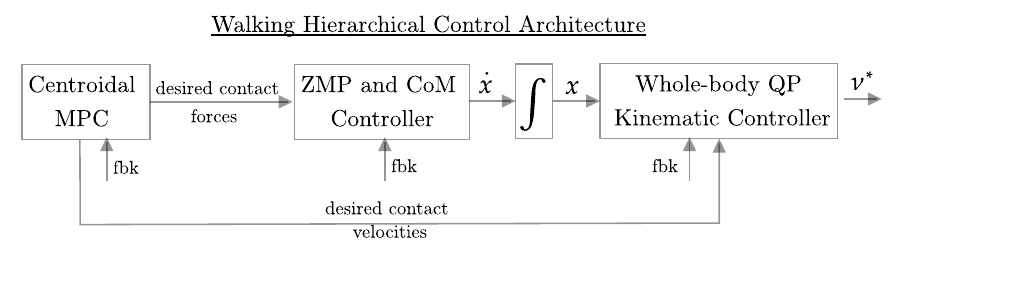}
\caption{The walking hierarchical control architecture, tuned via gradient-free techniques, composed of Centroidal Predictive Control (MPC) for calculating desired contact point forces and velocities, Zero Moment Point (ZMP) and Center of Mass (CoM) Controller for computing reference CoM velocity $\dot{x}$, and Whole-body Quadratic Programming (QP) Kinematic Controller that translates the previous block references into robots reference velocities $\nu^*$. Each layer processes feedback (fbk) from the robot's sensors.} 
\label{fig:herarchichal_architecture}
\end{figure*}
\noindent In this Section, we will briefly summarize the hierarchical control architecture utilized, which is visually depicted in Figure \ref{fig:herarchichal_architecture}. It is composed of three different layers:  Centroidal MPC, Zero Moment Point (ZMP) and Center of Mass (CoM) controller, and Whole-body Quadratic Programming (QP) kinematic controller. Each layer processes feedback from the robot's sensors and generates reference commands for the subsequent layers. In the following sections, we will briefly recall the foundations of each layer. We invite the interested reader to refer to the associated papers for more details.\looseness=-1
\subsubsection{Centroidal MPC}
\label{centroidal_MPC}
The centroidal MPC has been formulated analogously to the work done in  \cite{romualdi2022online}.
In the definition of the problem, we considered the contact locations $p_i$ as continuous variables with the following dynamics: $\dot{p}_i = (1-\gamma_i) v_i$, where $v_i$ is the contact velocity and $\gamma_i\in \left\{0,1\right\}$ is provided by a contact scheduler. The centroidal momentum is defined as   ${}_{G} h ^\top= \begin{bmatrix}
{}_{G} h^{p\top} & {}_{G} h^{\omega\top}
\end{bmatrix} \in \mathbb{R}^6$,  where ${}_{G} h^{p\top}$ and ${}_{G} h^{\omega\top}$ are respectively the aggregate linear and angular momentum of each link of the robot referred to the robot CoM. At instant $k$ and with a sampling period of $\Delta T$, one can define the following discretized dynamics: 
\begin{itemize}
\item Centroidal dynamics
\begin{equation}
    \label{dyn_1}
{}_G h[k+1] = {}_G h[k] + \Delta T\left( mg + \sum_{i = 1}^{n_c}\mathcal{P}_if_i\right),
\end{equation}
with $\mathcal{P}_i =\begin{bmatrix} \mathbb{I}_3 & 0_3 \\ (p_i - p_{CoM})^{\wedge} & \mathbb{I}_3 \end{bmatrix}.$
\item Contact dynamics: 
\begin{equation}
\label{dyn_2}    
p_i[k+1] =  p_i[k]  + \Delta T ((1 - \gamma_i) \ v_i).
\end{equation}
\item  CoM dynamics: 
\begin{equation}
\label{dyn_3}    
p_{CoM}[k+1] = p_{CoM}[k] + \Delta T\left(\frac{1}{m}C{}_G h[k]\right),
\end{equation}
where $C = \begin{bmatrix} \mathbb{I}_3 & 0_3\end{bmatrix}$ is a selector matrix.

\end{itemize}
Let, at time instant $k$, $\mathcal{X}_k ^\top = \begin{bmatrix} p_{\text{CoM}}[k]^\top & \prescript{}{G}{h}[k]^\top &  p_{i}[k]^\top \end{bmatrix}$ be the controller  state and
$\mathcal{U}_k ^\top = \begin{bmatrix} f_{i}[k]^\top &  v_{i}[k]^\top \end{bmatrix}$ the controller output. Moreover, let $\mathcal{K}_i$ identify the feasible region for the contact forces as in \cite{gabri_parametrization}. We define the following Optimal Control Problem (OCP): \looseness=-1
\begin{IEEEeqnarray}{RCL}
\label{centroidal_mpc} 
\IEEEyesnumber
& \minimize_{\substack{\mathcal{X}_k, \mathcal{U}_k,  \\  \nonumber k = [0, N]}}& \sum_{k = 0} ^ N   \left( \underset{k}{cost} \right), \quad \text{s.t.} \\ \nonumber
 &&\text{Centroidal discretized dynamics of } \eqref{dyn_1}\\ \nonumber
 && \text{Contact discretized dynamics of } \eqref{dyn_2}\\ \nonumber
 && \text{Center of mass discretized dynamics of } \eqref{dyn_3}\\ \nonumber
 && f_i \in  \mathcal{K}_i. \nonumber \\
\yesnumber
\end{IEEEeqnarray}
In such a formulation, the cost function is defined as: 
\begin{equation*}
    \nonumber
    \underset{k}{cost}=\sum_{i = 1}^{n_c} \left\lVert \Psi_{f_{i}} \right\lVert_{\bar{W}_f}^2 + \sum_{i = 1}^{n_c} \left\lVert\Psi_{\dot{f}_{i}}\right\lVert_{\bar{W}_{\dot{f}}}^2 +\Psi_{h} +  \sum_{i = 1}^{n_c} \left\lVert\Psi_{p_{i}}\right\lVert^2_{\bar{W}_{p_{i}}},
\end{equation*}
where $\Psi_{f_{i}}$ is a contact force regularization term, thus it drives the contact forces towards symmetric values; $\Psi_{\dot{f}_{i}}$ is a term aimed at reducing the rate of change of the contact forces; $\Psi_{h} = \left\lVert\Psi_{h^p}\right\lVert^2_{\bar{W}_{h^p}} + \left\lVert\Psi_{h^\omega}\right\lVert^2_{\bar{W}_{h^\omega}}$ is the centroidal momentum task, aimed at tracking a desired momentum trajectory, where  $\Psi_{h^p}$ and $\Psi_{h^\omega}$  are respectively the linear and angular part; $\Psi_{p_i}$ aims to regularize the contact location towards the nominal values. 
Finally, $\bar{W}_f,\bar{W}_{h^\omega},\bar{W}_{p_{i}},\bar{W}_{\dot{f}}$ and $\bar{W}_{h^p}$ are positive definite diagonal matrices.  \looseness=-1
\subsubsection{ZMP-CoM controller}
This layer computes the reference CoM velocity along the walking surface by approximating the motion of the humanoid robot through the 
\emph{Linear Inverted Pendulum Model} (LIPM), see \cite{Kajita2001}. The reference CoM velocity projected on the walking surface, denoted as $\dot{x}^*$, is defined by
\begin{equation}
\label{eq:zmp_controller}
\dot{x}^* = \dot{x}_{ref} - \bar{K}_{zmp}(r^{zmp}_{ref} - r^{zmp}) + \bar{K}_{com} (x_{ref} - x).
\end{equation}
In \eqref{eq:zmp_controller}, $x \in \mathbb{R}^2$ is the projection of the CoM position on the walking surface, while  $r^{zmp}\in \mathbb{R}^2$ is the position of the zero moment point (ZMP). The terms $r^{zmp}_{ref}$, $\dot{x}_{ref}$ and $x_{ref}$  are computed starting from the desired contact forces $f_i$ as outputs of the MPC block as in \cite{romualdi2022online}. %
Finally,  $\bar{K}_{com},\bar{K}_{zmp}  \in \mathbb{R}^{2\times 2}$ are diagonal matrices subject to the following constraints, deriving from the LIPM:  $\bar{K}_{com}> \omega \mathbb{I}_2$  and $0_2 < \bar{K}_{zmp} < \omega \mathbb{I}_2$, where $\omega$ is the inverse of the pendulum time constant; namely, denoting $z_0$ the CoM height, $\omega = \sqrt{g/z_0}$. 
\subsubsection{Whole-body QP kinematic controler}
\label{WholebodyKinematicController}
The whole-body QP kinematic controller has been implemented as in \cite{romualdi2020benchmarking} and it gives the reference robot velocity $\nu^*$ as output of the following optimization problem:

\begin{IEEEeqnarray}{RCL}
\label{ik}
\IEEEyesnumber
&\minimize_\nu & \left(  \Theta_\mathcal{T} + \Theta_s \right), \quad \text{s.t.}  \IEEEyessubnumber \label{cost_qp}\\ \nonumber
&&  J_{CoM} \nu = v^ * _ {{CoM}}   \IEEEyessubnumber \label{com_constraint} \\
&&  J_{\mathcal{F}} \nu = v^ * _ {\mathcal{F}}  \quad \forall \mathcal{F} \in \left \{\mathcal{RF},\mathcal{LF}\right\}  \IEEEyessubnumber \label{contact_constraint} 
\\ 
&&  \dot{s}^- \le \dot{s} \le \dot{s}^+. \IEEEyessubnumber \label{velocities_constraint}    
\end{IEEEeqnarray}
In the previous, \eqref{com_constraint} constraints the reference CoM velocity projected on the walking surface with  $v^*_{CoM} = \dot{x}^* - K^p_{CoM}(x - x^*)$, where $\dot{x}^*$ is computed by \eqref{eq:zmp_controller} while $x^*$ is its integral.
The constraint in \eqref{contact_constraint} forces the feet frame velocities to be equal to the reference velocities $v^*_ {\mathcal{F}}$, computed as:
\begin{equation*}
\label{feetVelocitiesStar}
v^*_ {\mathcal{F}} = {\dot{p}}^*_{\mathcal{F}} -
\begin{bmatrix}
K _{{\mathcal{F}}} \left( p_{ \mathcal{F}}-p^*_{ \mathcal{F}} \right)\\
K _{\omega  _{\mathcal{F}}} \text{log}({R}_{\mathcal{F}}{R} _{\mathcal{F}} ^{*^\top} )^{\vee} \qquad 
\end{bmatrix},
  \quad \forall \mathcal{F} \in \left \{ \mathcal{RF},\mathcal{LF}\right\}.
\end{equation*}
In the latter, $p^*_{F}, \dot{p}^*_{F}$ and $R^*_\mathcal{F}$ are computed starting from the contact point velocities $v_i$, output of the MPC, as in \cite{romualdi2020benchmarking}. By \eqref{velocities_constraint}, the joint velocities $\dot{s}$ are constrained by the lower and upper bounds $\dot{s}^-$ and $\dot{s}^+$. For what concerns the cost \eqref{cost_qp}, it is composed of two terms. The first one is a task driving the torso frame $\mathcal{T}$ towards desired orientation and position (along the z-axis only) and is defined as 
$\Theta_\mathcal{T}=\frac{1}{2} \left\lVert v^* _ {\mathcal{T}}-J_{\mathcal{T}} \nu \right\lVert^2_{ K_{\mathcal{T}}}$,
with  $K_{\mathcal{T}}$ positive definite   and $v^*_{\mathcal{T}}$ computed as
\begin{equation}
\label{torsoVelocitiesStar}
v^*_{\mathcal{T}}=\begin{bmatrix}
    \dot{p}_{z,\mathcal{T}}^* \\
    \omega_\mathcal{T}^*\\
\end{bmatrix} = \begin{bmatrix}
    \dot{p}_{z,\mathcal{T}}^d \\
    \omega_\mathcal{T}^d\\
\end{bmatrix} -
\begin{bmatrix}
K_{z_\mathcal{T}} \left( p_{\mathcal{T},z}-p^*_{ \mathcal{T},z} \right)\\
K _{\omega _{\mathcal{T}}} \text{log}({R}_{\mathcal{T}}{R} _{\mathcal{T}} ^{*^\top} )^{\vee} \qquad 
\end{bmatrix}.
\end{equation}
The second one is given by $\Theta_s=\left\lVert \dot{s} - \dot{s}^*\right\lVert^2_\Lambda$ and is the postural task,  where $\Lambda$ is a given positive definite matrix. It promotes a reference joints velocity  $\dot{s}^ * = -K_{s} (s - s^d)$, where $K_{s}$ is a given positive definite matrix. . 
Finally, the output $\nu^*$ is then integrated, and the reference joint position $q$ is given as input to the robot's low level.  

\section{Parameters Optimization}
\label{sec:gain_optimization}
\noindent Given the previously defined hierarchical control architecture, as illustrated in the diagram of Figure \ref{fig:herarchichal_architecture}, we aim to identify the optimal gains and weights $\xi \in \Xi$ characterizing the layers of the hierarchical control architecture that allow solving the walking task. 
Formally, we want to find the parameters (i.e. gains and weights) $\xi^*$  such that
\begin{equation}\label{gain_tuning}
    \xi ^* \in \argmax\limits_{\xi \in \Xi} \mathcal{G}(\xi).
\end{equation}
With  $\mathcal{G}: \Xi \to \mathbb{R}$ an {\it objective} function designed to measure the quality of a given parameter configuration in solving a walking task.  
Given a parameter vector $\xi \in \Xi$, the associated objective function value is computed by executing the hierarchical control architecture employing the entries of $\xi$ as gains and weights for the different layers.
Notice that several configurations might be unfeasible, meaning they do not solve the problem and may cause the robot to fall, leading to potential damage. Since we assume no initial feasible configuration is provided, to avoid the risk of damaging a real robot, the evaluation of the function $\mathcal{G}$ must rely on a simulator. Then, due to this dependence on the simulator, the gradient of the objective function is not accessible or even defined and, thus, optimization methods that do not rely on gradients have to be used.

\noindent In the next sections, we describe how we tackle this problem. We first define the search space and describe which parameters we optimize (Sec. \ref{sec:search_space}). Then we propose an objective function that relates the quality of a configuration with the simulation time. We finally extend our proposal allowing us to consider feasible solutions that minimize the mean torque (Sec. \ref{sec:fitness_definition}).
\subsection{Search Space and Parameters}
\label{sec:search_space}
\noindent We define the parameter vector $\xi \in \Xi \subseteq \mathbb{R}^{14}$ as the concatenation of the weights characterizing the centroidal MPC layer $\xi^{MPC} \in \Xi_{MPC} \subset \mathbb{R}^7$, the feedback gains characterizing the ZMP-CoM controller $\xi^{ZMP} \in \Xi_{ZMP} \subset \mathbb{R}^2$ and the whole body QP controller $\xi^{QP} \in \Xi_{QP} \subset \mathbb{R}^5$; i.e. 
\begin{equation}
\label{eq:construct_x}
    \xi \coloneqq\begin{bmatrix}
        \xi^{MPC}, \xi^{ZMP}, \xi^{QP}
    \end{bmatrix}.
\end{equation}
The search space $\Xi = \Xi_{MPC} \times \Xi_{ZMP} \times \Xi_{QP}$ is thus a subspace of $\mathbb{R}^{14}$ and it is fixed a priori, based on physical constraints such as the LIPM model constraint and positive definiteness.
In the next paragraphs, we describe these three components.

\paragraph*{The weights of the centroidal MPC layer}
They are the weights of the cost function in \eqref{centroidal_mpc}, 
namely $\bar{W}_f, \bar{W}_{p_{i}},\bar{W}_{h^\omega}\bar{W}_{\dot{f}}, \bar{W}_{h^p} \in \mathbb{R}^{3 \times 3}$. 
For simplicity, we assume the weights are structured as $\bar{W}_{\#} = W_{\#} \mathbb{I}^3$, where $W_{\#} \in \mathbb{R}$, except for $\bar{W}_{\dot{f}}$ and $\bar{W}_{h^p}$, which are supposed to be diagonal matrices and so represented by their diagonal elements $W_{\dot{f}}, W_{h^p} \in \mathbb{R}^{3}$.
Additionally, by enforcing the weight along the $x$-axis to be equal to the weight along the $y$-axis, to enforce symmetrical robot behavior, we can fully describe the cost function weights with the following vector:
\begin{equation*}
    \xi^{MPC}\coloneqq \begin{bmatrix}
        W_f, W_{\dot{f}_{xy}},W_{\dot{f}_z}, W_{h^p_{xy}},W_{h^p_{z}}, W_{h^\omega},W_{p_{i}}
    \end{bmatrix}.
\end{equation*}

\paragraph*{Feedback gains of the ZMP-CoM controller} 
The ZMP-CoM control law is characterized by the gains, two diagonal matrices $\bar{K}_{zmp}$ and $\bar{K}_{com} \in \mathbb{R}^{2 \times 2}$. In this case, we will refer to $K_{zmp}$ and $K_{com} \in \mathbb{R}^2$ as the diagonal elements of the matrices $\bar{K}_{zmp}, \bar{K}_{com}$ respectively. Again, by enforcing that the $x$-axis component is equal to the $y$-axis component, we can define $\xi^{ZMP} \in \mathbb{R}^2$ as:

    \begin{equation*}
        \xi^{ZMP} \coloneqq \begin{bmatrix}
        K_{zmp}, K_{com}     \end{bmatrix}.
    \end{equation*}

\paragraph*{The whole body QP controller parameters} For the whole body QP problem, several gains can be tuned between the cost function and the constraints of \eqref{ik}. We choose to consider the following matrices: $K_{\mathcal{F}}, K_{\omega \mathcal{F}}, K^p_{CoM}, K_{z, \mathcal{T}}, K_{\omega, \mathcal{T}} \in \mathbb{R}^{3 \times 3}$. For simplicity, we will assume that the gains are in the form $K_{\#} = k_{\#} \mathbb{I}_3$, with $k_{\#} \in \mathbb{R}$. We will then define the set of gains that characterize the QP problem as: 
\begin{equation*}
    \xi^{QP} \coloneqq \begin{bmatrix}
        k_{\mathcal{F}},k_{\omega \mathcal{F}}, k^p_{CoM},k_{z, \mathcal{T}},k_{\omega, \mathcal{T}}
    \end{bmatrix}.
\end{equation*}
For the whole body QP controller, we will consider all the feedback gains except the joint regularization one. This is because the joint regularization has a high dimensionality ($n$) but is used only as a regularizer, while the other parameters ensure task accomplishment.

\subsection{Objective functions}
\label{sec:fitness_definition}
\noindent In this Section, we define two different objective functions to measure the quality of parameter configurations $\xi$. The first objective function $\mathcal{G}_1$ evaluates the quality of a parameter configuration $\xi$ measuring the duration of the time the robot can walk. Let $t : \Xi \to \mathbb{R}$ be the function that takes a parameter configuration $\xi$ and returns the time the robot walked without falling using $\xi$ as gains and weights and let $t^*$ be the nominal trajectory execution time. We define the first target function $\mathcal{G}_1$ as: \looseness -1
\begin{equation}\label{eqn:target_g1}
    \mathcal{G}_1(\xi) := \left[ t^* - t(\xi) \right]^{-1}.
\end{equation}
Notice that  $t(\xi) \approx t^*$ holds if the robot performs the whole trajectory without falling, thus, a solution $\xi^*$ that maximizes \eqref{eqn:target_g1} allows the robot to walk for the entire trajectory. 
The intuition behind this obpreservedjective function is that similar parameter configurations should yield similar function values. Consequently, unfeasible configurations that allow the robot to perform many steps (i.e., high $t(\xi)$) should be close to feasible configurations. 
Moreover, notice that this target function effectively distinguishes between unfeasible configurations that do not permit any step and those that allow many steps. This characteristic should help optimization algorithms explore the parameter space by indicating which configurations are better, thereby guiding the search toward feasible solutions more efficiently. This formulation can be extended by defining another objective function, $\mathcal{G}_2$, which also takes into account $\tilde{\tau}(\xi)$, the mean torque of the joints obtained using the parameters $\xi$. Given $W_1, W_2 \geq 0$, we define \looseness=-1
\begin{equation}\label{eqn:target_g2}
    \mathcal{G}_2(\xi) :=   \left[ W_1 \left( t^* - t(\xi) \right) + W_2\left\lVert \Tilde{\tau}(\xi) \right\rVert \right]^{-1}.
\end{equation}
Maximizing $\mathcal{G}_2$ permits to find parameters $\xi$ that allow the robot to perform the whole trajectory without falling while minimizing the norm of the mean joint torques. Notice that $\mathcal{G}_1$ is a special case of $\mathcal{G}_2$ with  
$W_1 = 1$ and $W_2 = 0$. 

\begin{figure*}{}
\centering
\begin{subfigure}{\columnwidth}
    \centering
    \includegraphics[trim=0.0cm 0.0cm 3.2cm 0.0cm, clip=true, width=\columnwidth]{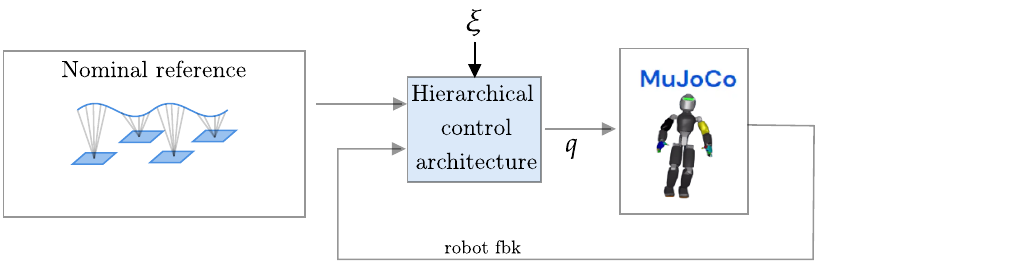}
    \caption{}
    \label{fig:fitness_computation}
\end{subfigure}
\hfill
\begin{subfigure}{\columnwidth}
    \centering
    \includegraphics[trim=0.0cm 0.0cm 3.2cm 0.0cm, clip=true, width=\columnwidth]{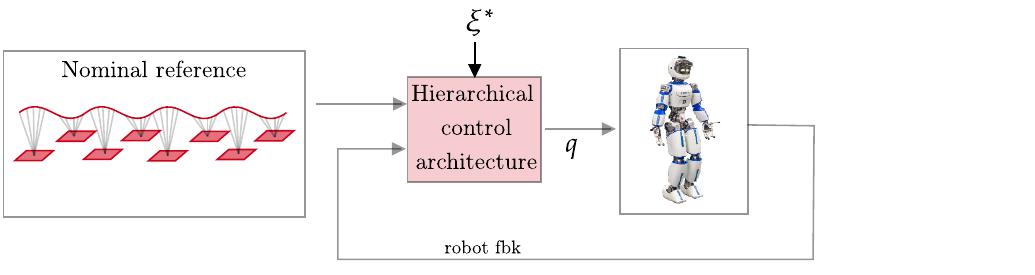}
    \caption{}
    \label{fig:validation}
\end{subfigure}
 \hfill
\caption{(a) The infrastructure used to evaluate the objective function during optimization, leveraging the MuJoCo simulator to assess robot behavior with a hierarchical control architecture defined by the parameter set $\xi$. (b) The infrastructure employed to test the optimal gain set $\xi^*$ on the real ergoCub robot. Note that the nominal reference to evaluate the objective function differs from the reference used for validating the optimal parameters on the real robot.}
\label{fig:optimizationAndTesting}
\end{figure*}
\section{RESULTS}
\label{sec:results}
\noindent To identify and validate the optimal parameter configurations, 
we implemented two distinct infrastructures, depicted in Figure \ref{fig:optimizationAndTesting}. The first infrastructure, in Figure \ref{fig:optimizationAndTesting}(a),  utilizes the MuJoCo simulation environment \cite{todorov2012mujoco} and it is used to optimize the objective functions. The second infrastructure, in Figure \ref{fig:optimizationAndTesting}(b), is used for validation and involves testing the identified optimal parameters $\xi^*$ on the real ergoCub robot. In both cases, the control architecture implementation was based on \texttt{bipedal-locomotion-framework}\footnote{\url{github.com/ami-iit/bipedal-locomotion-framework}}.

\noindent In Sec. \ref{sec:training}, we present the results of the parameter optimization process, performed with different gradient-free optimization techniques. 
Following that, in Sec. \ref{sec:validation}, we discuss the results of the real robot experiments.

\subsection{Parameters Optimization}
\label{sec:training}

\begin{table}[]
\begin{tabular}{c|c|c}
    &Lower Limit & Upper Limit \\ \hline
   \cellcolor{cyan!25}$\Xi_{MPC}$ 
& $\left[ 10,10,10, 2, 80, 10, 10\right]$ &$\left[ 150,150,150,50, 140,150,150\right]$ \\ 
\cellcolor{cyan!25}$\Xi_{ZMP}$& $\left[ 0.5, 3.5\right]$&  $\left[ 1.0,5.0\right]$\\ 
\cellcolor{cyan!25}$\Xi_{QP}$& $\left[2.5, 1.0, 1.0,1.0,1.0\right]$&  $\left[5.0,10.0,5.0,5.0,10.0\right]$\\ \hline
\end{tabular}
\caption{Hierarchical architecture parameter search space. }
\label{tab:search_space}
\end{table}

\begin{figure*}
\centering
\begin{subfigure}{\columnwidth}
    \includegraphics[trim=0.5cm 0.0cm 0.5cm 0.0cm, clip=true, width=\textwidth]{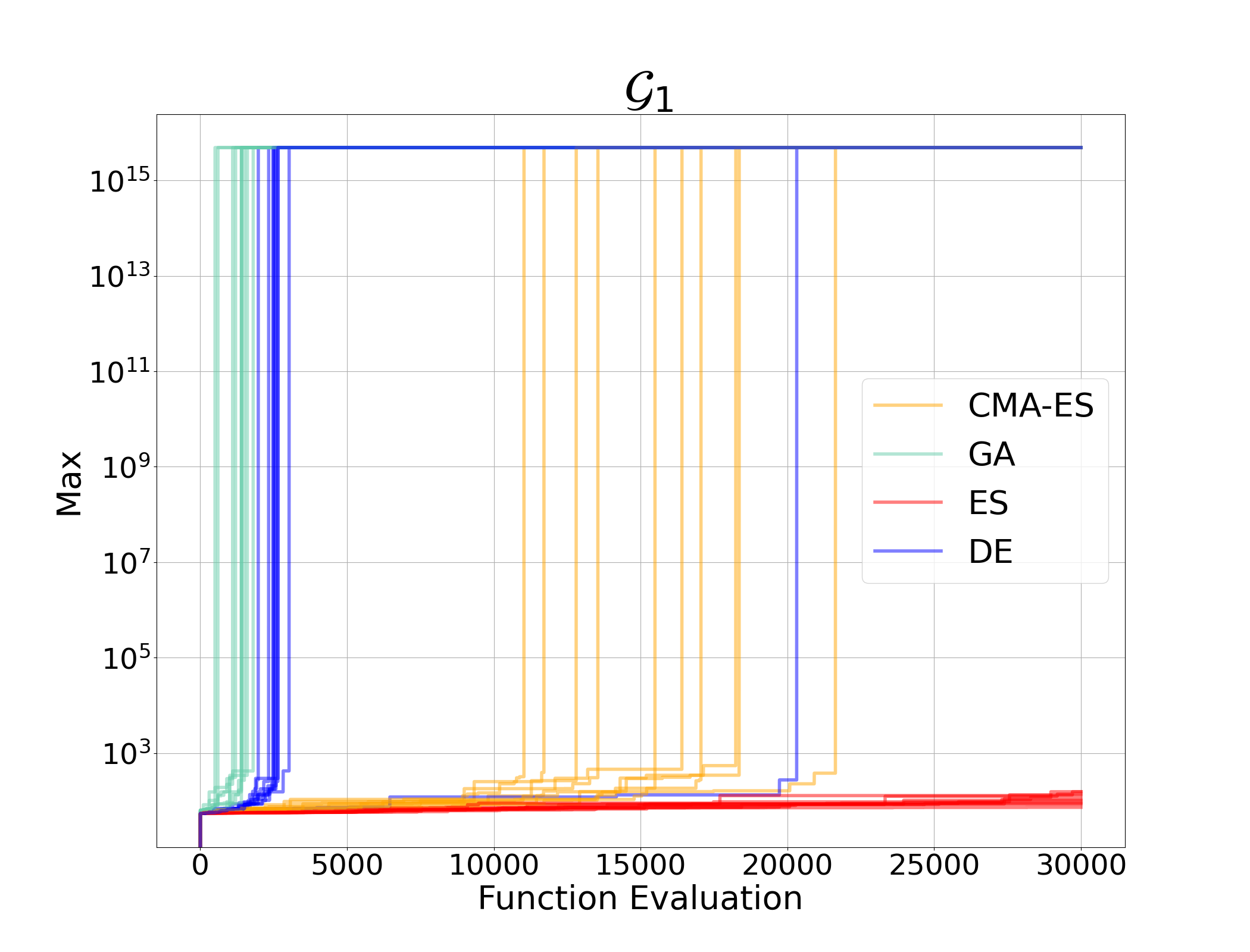}
    \caption{}
    \label{fig:first}
\end{subfigure}
\hfill
\begin{subfigure}{\columnwidth}
    \includegraphics[trim=0.5cm 0.0cm 0.5cm 0.0cm, clip=true, width=\textwidth]{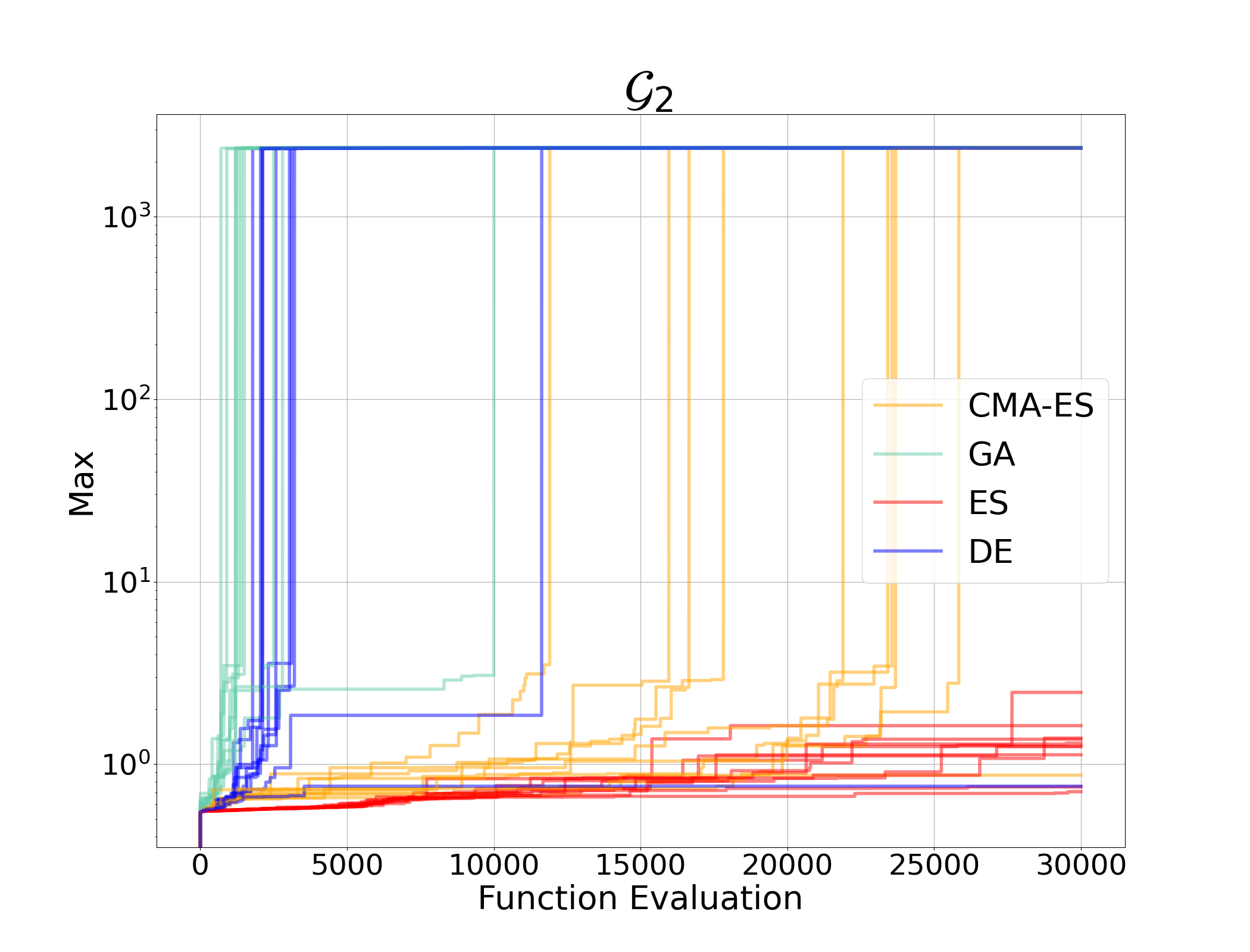}
    \caption{}
    \label{fig:second}
\end{subfigure}
 \hfill
 \begin{subfigure}{\columnwidth}
\centering
 \begin{tabular}{|c|c|c|c|c|} \hline
& \cellcolor{cyan!25}GA &\cellcolor{cyan!25} CMA-ES &\cellcolor{cyan!25} DE & \cellcolor{cyan!25} ES  \\ \hline
$\mu$ &   $4.85\times 10^{15}$ &$4.85\times 10^{15}$ & $4.85\times 10^{15}$ & $1.13 \times 10^{2}$ \\\hline
$\sigma$ & $0.0$&$0.0$ &$0.0$ &$28$ \\ \hline
\end{tabular}
\caption{$\max \left( \mathcal{G}_1\right)$}
\label{tab:G_2}
\end{subfigure}
\hfill
 \begin{subfigure}{\columnwidth}
\centering
 \begin{tabular}{|c|c|c|c|c|c|} \hline
& \cellcolor{cyan!25}GA &\cellcolor{cyan!25} CMA-ES &\cellcolor{cyan!25} DE &\cellcolor{cyan!25} ES \\ \hline
$\mu$& $2.386 \times 10^3$ &$2.145 \times 10^3$  &$2.148 \times 10^3$  &$1.32$   \\\hline
$\sigma$ & $4.04$  & $714.76$& $715.86$ &$0.46$ \\ \hline
\end{tabular}
\caption{$\max \left( \mathcal{G}_2 \right)$}
\label{tab:G_1}
\end{subfigure}
\hfill
\caption{The performances of the different gradient-free optimizers based on $10$ independent runs. The y-axis represents the maximum objective function value achieved, plotted on a logarithmic scale. On the x-axis is the number of function evaluations performed.  Each line corresponds to a different run of the algorithm. In (a) the objective value was defined as \eqref{eqn:target_g1} while in (b) it was defined as \eqref{eqn:target_g2}. In (c) and (d) the mean $\mu$ and standard deviation $\sigma$ of the maximum objective function value found in the different runs.}
\label{fig:training}
\end{figure*}
\noindent To optimize the parameters of the infrastructure, we compare different gradient-free optimization techniques, namely Covariance Matrix Adaptation Evolution Strategy (CMA-ES) \cite{Hansen2006}, Genetic Algorithm (GA) \cite{Whitley1994},  Evolution Strategy (ES)  \cite{nevergrad}, and Differential Evolution (DE) \cite{BILAL2020103479}. 
We optimize the two different objective functions, $\mathcal{G}_1$ and $\mathcal{G}_2$, and we use the architecture illustrated in Figure \ref{fig:optimizationAndTesting}(a). 
Such a hierarchical architecture controls the robot in MuJoCo to perform a forward walking nominal trajectory that lasts $t^* = 20$s. The search space $\Xi$, summarized in Table \ref{tab:search_space}, is defined based on the constraints identified in Sec. \ref{sec:background} (e.g., the LIPM constraints) and the physical significance of the parameters. For what concerns the weights of the objective function of \eqref{eqn:target_g2}, we set $W_1 = 100$ and $W_2= 0.001$.   
For each algorithm and target function, we performed $10$ independent runs, each initialized randomly, for a total of $80$ independent runs. We fixed a budget of $30 \times 10^3$ function evaluations for every optimization algorithm. The experiments have been performed on a machine with the following specifications: \texttt{CPU: AMD EPYC 7513 32-Core Processor @ 2.60GHz ×4, RAM: 1024GB DDR4-3200}. The GA was implemented using the \texttt{pygad} library \cite{gad2021pygad}, with
the \texttt{k-tournament} \cite{goldberg1991comparative} selection method (with $k=4$) and \texttt{two-point} crossover. The mutation is random with a $10$\% gene mutation probability while the elitism parameter is set to $10$. The other algorithms are implemented using \texttt{Nevergrad} \cite{nevergrad}. The CMA-ES algorithm is initialized by setting the mean of the multivariate Gaussian to the centroid of the search space $\Xi$ and the covariance to $\sigma^2 I$, with $\sigma^2 = 10.0$. For the DE algorithm, the crossover rate is set to $0.5$ and the differential weights are set to $0.8$. For the ES algorithm, the recombination rate is set to $1.0$ and the offspring size is set to $99$. The population size for each method is set to $100$. Figures \ref{fig:first} and \ref{fig:second} show the best values found optimizing $\mathcal{G}_1$ and $\mathcal{G}_2$, respectively, against the number of function evaluations for all the algorithms across all runs. Tables \ref{tab:G_1} and \ref{tab:G_2} summarize the mean and standard deviation of the objective values at the optimal configuration 
for all $10$ independent runs per algorithm. 
When optimizing $\mathcal{G}_1$, all algorithms except ES find a feasible configuration that enables the robot to walk. GA converges to the optimal value in at most $15 \times 10^2$ function evaluations, while DE requires a comparable amount of resources, and CMA-ES requires $11 \times 10^3$ function evaluations. When optimizing $\mathcal{G}_2$, GA remains the fastest algorithm, converging within $10 \times 10^3$ function evaluations in the worst-case scenario, whereas DE and CMA-ES require up to $20 \times 10^3$ function evaluations. DE and CMA-ES also fail to converge within the given budget in one out of the ten runs analyzed. Additionally, the mean value of the maximum found with GA across different runs is $2.386 \times 10^3$, surpassing the results achieved by the other algorithms. GA exhibits a smaller standard deviation of $4.04$ compared to CMA-ES and DE, which have standard deviations around $700$, primarily due to their failure cases. ES fails to converge within the allocated budget in this case as well. Therefore, the results indicate that GA requires fewer function evaluations to identify optimal parameter configurations and achieves a $100$\% success rate. 
We observe that there are significant variations in the solutions found  
by the different employed algorithms. This suggests  
that different parameter configurations $\xi$ can enable the robot to walk. It is important to note that since the target function is not concave, the convergence to an optimal configuration is not guaranteed. Indeed, as seen with ES, not every algorithm solves the problem. 
When optimizing $\mathcal{G}_2$, which involves minimizing the robot joint torques, some solutions have higher contact force symmetry weights $W_f$  
compared to those found when optimizing $\mathcal{G}_1$. A similar trend is observed for the ZMP gains $K_{zmp}$. This suggests that solutions ensuring symmetric values on the wrenches and better tracking of the ZMP result in lower required torque. However, the high variation in the solutions found prevents drawing definitive conclusions. Adding more constraints could guide the algorithms toward more consistent and unified solutions.\begin{figure*}
\centering
\begin{myframe}{Validation on real robot}
\begin{subfigure}{1.0\textwidth}
    \includegraphics[width=\textwidth]{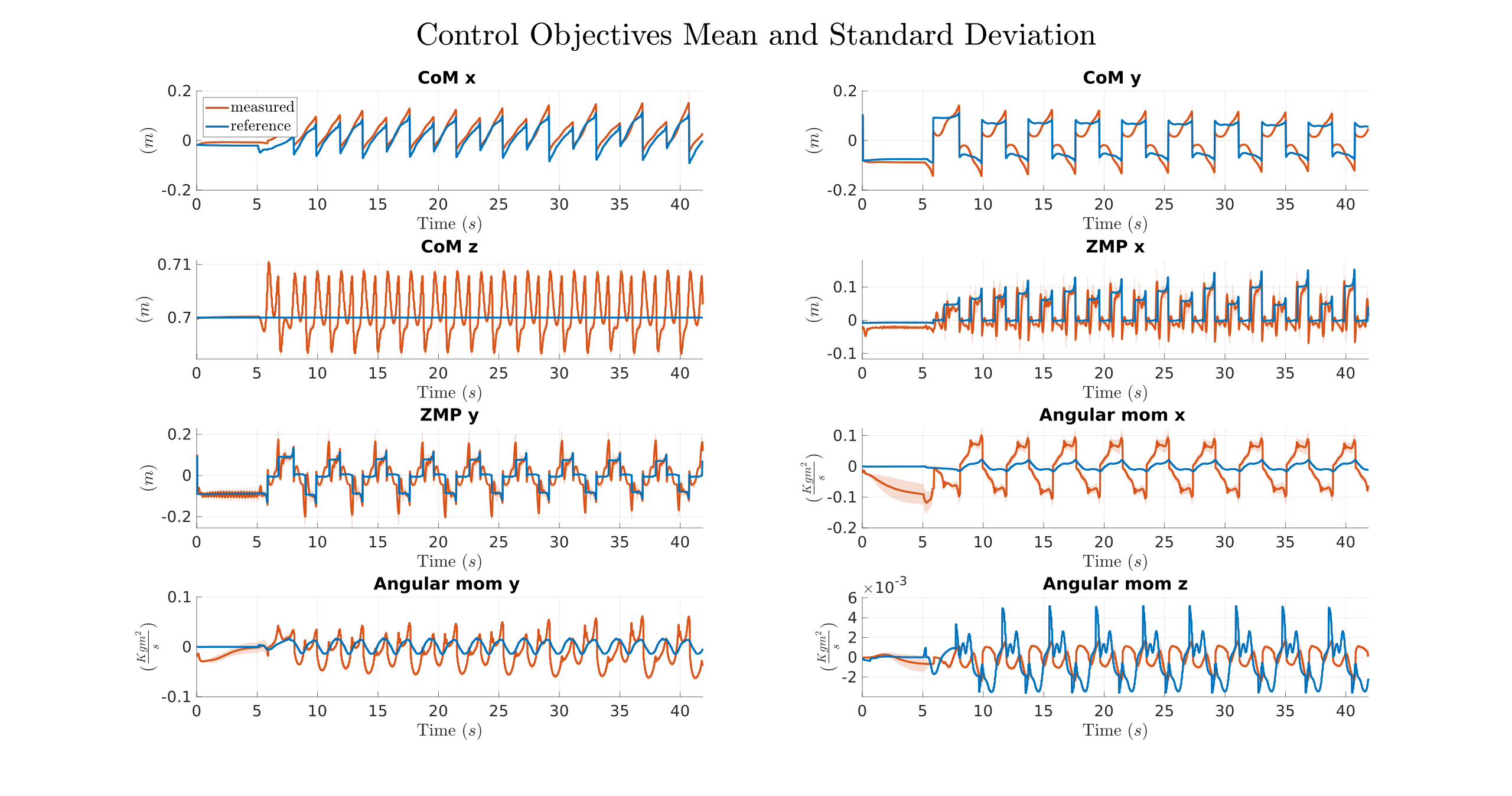}
    \caption{}
    \label{fig:real_robot}
\end{subfigure}
\hfill
\centering
\begin{subfigure}{1.0\textwidth}
   \includegraphics[width=\textwidth]{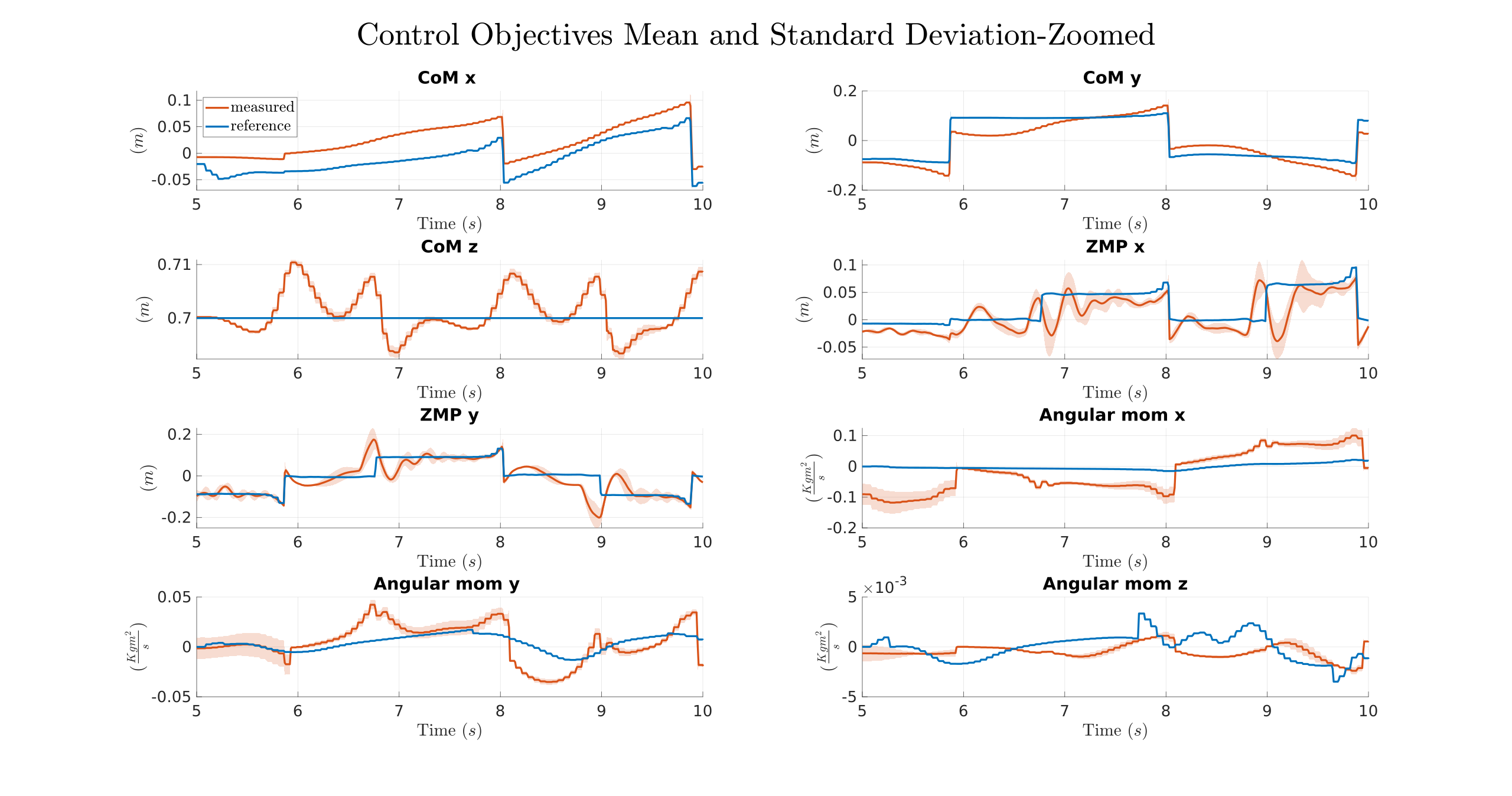}
    \caption{}
    \label{fig:real_robot_zoomed}
\end{subfigure}
\end{myframe}
 \hfill
\hfill
\caption{Mean and standard deviation of control objectives over $20$ experimental runs on the real robot using the optimal parameters identified from $20$ independent GA runs. (a) Trajectories of the CoM, ZMP, and angular momentum over the entire validation trajectory. (b) A zoomed-in segment of the trajectories.}
\end{figure*}
\subsection{Real Robot Validation}
\label{sec:validation}
\noindent To validate the proposed approach, we employ the architecture of Figure \ref{fig:optimizationAndTesting}(b) to control the real humanoid robot ergoCub to perform a walking task. The optimal gains previously found optimizing $\mathcal{G}_1$ (and $\mathcal{G}_2$) are used. The reference trajectory used during validation differs from that used in the optimization phase, as demonstrated in the attached video. Specifically, the validation trajectory involves walking in a parabolic path with swinging arms, whereas the training phase focuses solely on forward walking.
We initially conducted a manual trial to tune the parameters by hand. With these initial gains, the robot was able to take only a few steps, as depicted in the accompanying video, demonstrating that not all parameter configurations led to successful walking. We subsequently tested the optimized parameters identified by GA, which obtained the best performances during the optimization phase. In total, we tested $20$ different configurations on the real robot. In Figure \ref{fig:real_robot}, we present the mean and standard deviation of the measured and reference trajectories related to various control objectives. Specifically, we include trajectories for the CoM, ZMP, and angular momentum. Additionally, Figure \ref{fig:real_robot_zoomed} provides a detailed $5$-second zoom-in of these trajectories. The tests showed that all configurations found in the optimization phase enabled the robot to complete the entire trajectory, thus achieving a $100$\% success rate on the real robot.  Furthermore, despite the variations in the optimal parameters across different runs, the performances were comparable, as indicated by the small standard deviation observed in the trajectories in Figure \ref{fig:real_robot_zoomed}. Nevertheless, the robot behavior slightly changes with the different configurations, as highlighted in  Figure \ref{fig:real_robot_zoomed}, particularly noticeable in the larger standard deviation of the ZMP. This variability could be attributed to the previously identified disparities in the optimized parameters for $\mathcal{G}_1$ and $\mathcal{G}_2$, specifically in terms of identified value for the ZMP gain.
\section{CONCLUSIONS}
\label{sec:conclusions}
\noindent This paper introduces a framework for automatic tuning of a complete cascade control architecture using gradient-free techniques. We compare different black-box optimization algorithms to maximize the objective. In particular, GA show the fastest convergence, requiring a maximum of $10 \times 10^3$ function evaluations. 
The optimized parameters are successfully transferred to the real ergoCub robot for performing a walking trajectory different from the one considered in the optimization process.  However, the variability in optimal parameter solutions highlights a need for further refinement. Future work could focus on including additional constraints 
to improve performance, such as walking speed or energy efficiency and extending the 
comparison to other zeroth order optimization algorithms like \cite{NEURIPS2023_7429f4c1,sszd}.

\paragraph*{Acknowledgements} {\small The paper was supported by the Italian National Institute for Insurance against Accidents at Work (INAIL) ergoCub Project. M.R., C.M., and L.R. acknowledge the financial support of the European Research Council (grant SLING 819789), the European Commission (ELIAS 101120237), the US Air Force Office of Scientific Research (FA8655-22-1-7034), the Ministry of Education, University and Research (grant ML4IP R205T7J2KP; grant BAC FAIR PE00000013 funded by the EU - NGEU) and the Center for Brains, Minds and Machines (CBMM), funded by NSF STC award CCF-1231216. M.R. and C.M. are members of the Gruppo Nazionale per l’Analisi Matematica, la Probabilità e le loro Applicazioni (GNAMPA) of the Istituto Nazionale di Alta Matematica (INdAM). The European Commission and the other organizations are not responsible for any use that may be made of the information it contains. This work represents only the view of the authors.}\looseness=-1

\bibliographystyle{IEEEtran}
\bibliography{bibliography}

\end{document}